\documentclass{article}

\usepackage{graphicx}
\usepackage{comment}
\usepackage{amsmath,amssymb} 
\usepackage{color}

\newcommand{\rot}{r} %
\newcommand{\ES}{3D-OES} %

\newcommand{\Enc}{E_{\mathrm{GRNN}}}

\newcommand{\Det}{Det}

\newcommand{\Btwo}{\textit{graph-XYZ-image}}
\newcommand{\Bone}{\textit{graph-XYZ}}
\newcommand{\VF}{\textit{VF}}
\newcommand{\Pla}{\textit{PlaNet}}

\newcommand{\M}{\textbf{M}} %
\newcommand{\x}{\textbf{x}} 

\newcommand\todo[1]{\textcolor{red}{#1}}

\usepackage[final]{corl_2020} 

\usepackage{lipsum}

\newcommand\blfootnote[1]{%
  \begingroup
  \renewcommand\thefootnote{}\footnote{#1}%
  \addtocounter{footnote}{-1}%
  \endgroup
}

\renewcommand\footnotemark{}

\title{3D-OES: Viewpoint-Invariant Object-Factorized  Environment Simulators}

\author{Hsiao-Yu Fish Tung$^*$, Zhou Xian$^*$, Mihir Prabhudesai, Shamit Lal, Katerina Fragkiadaki\\
Carnegie Mellon University \\
\texttt{\{htung, xianz1, mprabhud, shamitl, katef\}@cs.cmu.edu}\\
\text{$^*$ These authors contributed equally to this work.}
}

\begin{document}
\maketitle


\begin{abstract}
We propose an  action-conditioned  dynamics model that predicts scene  changes caused by object and agent interactions in a viewpoint-invariant 3D neural scene representation space, inferred from RGB-D videos. 
In this 3D feature space, objects do not interfere with one another and their appearance persists over time and across viewpoints.  This permits our model to predict  future scenes long in the future by simply ``moving"  3D object features  based on cumulative object  motion predictions. Object motion predictions are computed by a graph neural network that operates over the object features extracted from the 3D neural scene representation. 
 Our model's simulations can be decoded by a neural renderer into 2D image views from any desired viewpoint, which aids the interpretability of our latent 3D  simulation space. 
 We  show  our model   generalizes well  its predictions across  varying number and appearances of interacting objects as well as across camera viewpoints, outperforming existing 2D and 3D dynamics models.   
We further demonstrate sim-to-real transfer of the learnt dynamics by applying our model trained solely in simulation to model-based  control for pushing objects to desired locations under clutter on a real robotic setup. 
\end{abstract}

\noindent \blfootnote{\hspace{-18pt}  Videos and source code are available at \href{https://zhouxian.github.io/3d-oes/}{https://zhouxian.github.io/3d-oes/}. }
\section{Introduction} \label{sec:intro}

Humans can effortlessly imagine how a scene will change as a result of their interactions with the objects in the scene \cite{doi:10.1080/03004430.2012.681649,Decety1990BrainSP}. 
What is the representation space of these imaginations? They are not pixel accurate and, interestingly, they are not affected by occlusions.   
Consider  a teaspoon dipping inside a coffee mug. 
Though it will be occluded from nearly all viewpoints but the bird's eye view, we have no difficulty keeping it in our mind as present and complete.  We can  imagine watching it from different viewpoints, increase or decrease its size, predict whether it will fit inside the mug, or even imagine filling the mug with more spoons.


Inspired by human’s capability to simulate scene changes in a viewpoint-invariant and occlusion-resistant manner, we  present 3D object-factorized environment simulators (\ES), an  action-conditioned  dynamics model that predicts scene  changes caused by object and agent interactions in a viewpoint-invariant 3D neural scene representation space, inferred from RGB-D videos.  
\ES \  differentiably maps an RGB-D image to a 3D  neural scene representation, detects objects in it, and forecasts their future 3D motions, conditioned on actions of the agent. 
A graph neural network operates on the extracted 3D object feature maps and the action input and predicts object 3D translations and rotations. Our model then generates future 3D scenes  by simply translating and rotating object 3D feature maps, inferred from the \textit{first time step}, according to cumulative 3D motion predictions. In this way, we avoid distribution shift in object features caused by forward model unrolling, hence minimizing error accumulation.


  Our main insight  is that scene dynamics are simpler to learn and represent in 3D  than in 2D, for the following  reasons:  
i) \textbf{In 3D, object appearance and object location are disentangled.} 
This means object appearance (what) does not vary  with  object locations (where). 
 This what-where disentanglement permits  generating scene  variations by simply translating and rotating  3D object appearance representations. 
 Scene generation by moving around objects is not possible in a projective 2D image space, since objects change appearance due to camera viewpoint variation, occlusions or out-of-plane object rotations  \cite{DBLP:journals/corr/abs-1710-05268}. It is precisely the permanence of object appearance in 3D that permits easy simulation. 
ii) \textbf{In 3D, inferring free space and object collisions  is easy.}  Given a 3D scene description in terms of object locations and 3D shapes, we can easily predict whether an object will collide with another or will be contained in another. Similar  inferences would require  many examples to learn directly from 2D images, and would likely have poor generalization. 
Yet, extracting  3D scene representations from  RGB or RGB-D video streams  is   a challenging  open problem in computer vision research   \cite{kulkarni2015deep,romaszko2017vision,commonsense,DBLP:journals/corr/IzadiniaSS16,Romaszko_2017_ICCV}. 
We build upon the recently proposed  geometry-aware recurrent neural networks (GRNNs) \cite{commonsense,adam3d} to infer 3D scene feature  maps from RGB-D images in a differentiable manner,  optimized end-to-end for our object dynamics  prediction task. 

We evaluate \ES \  in single-step and multi-step object motion prediction for object pushing and falling, and apply it for planning to push objects to desired locations. We test its generalization while varying the number and appearance of objects in the scene, and the camera viewpoint. 
We compare against existing learning-based 2D image-centric or object-centric  models of dynamics \cite{hafner2019planet, ye2019ocmpc} as well as graph-based dynamics learned over engineered 3D representations of object locations \cite{pmlr-v80-sanchez-gonzalez18a}. 
Our model outperforms them by a large margin. 
In addition, we empirically show that training the 2D baselines under varying viewpoints causes them to dramatically underfit on the training data, and be highly inaccurate in the validation set. This suggests that different architectures are necessary to handle viewpoint variations in dynamics learning and \ES~  is one step in that direction.

  In summary, the main contribution of this work is a  graph neural network over  3D object feature maps extracted from convolutional end-to-end differentiable 3D neural scene representations for forecasting 3D  object motion. Graph networks are widely used in 2D object motion interaction predictions \cite{DBLP:journals/corr/abs-1812-10972,ye2019ocmpc,DBLP:journals/corr/BattagliaPLRK16,kipf2018neural}. We show that by porting such relational reasoning in an  3D object-factorized space,  object motion prediction can  generalize across camera viewpoints, lifting a major limitation of previous works on 2D object dynamics. Moreover, future and counterfactual scenes can be easily generated by translating and rotating 3D object feature representations. In comparison to recent 3D particle graph networks \cite{particlegt,particlegt2,particledense}, our work can operate over input RGB-D images and not ground-truth particle graphs. In comparison to recent scene-specific image-to-3D particle graph encoders \citep{particlevisual}, our image to 3D scene encoder can generalize across environments with novel objects, novel number of objects, and novel camera viewpoints, Moreover, our model presents effective sim-to-real transfer to a real-world robotic setup.


\vspace{-0.1in}
\section{Related  Work} \label{sec:related}
\vspace{-0.1in}
The inability of systems of physics equations to capture the complexity of the world  \cite{NIPS2015_5780} has led many researchers to pursue learning-based models of dynamics, or combine those with analytic physics models to help fight the undermodeling and uncertainty of the world \cite{DBLP:journals/corr/abs-1903-11239,DBLP:journals/corr/abs-1808-03246}.  
Learning world models is both useful for model-based control \citep{hafner2019planet,DBLP:journals/corr/abs-1803-10122,DBLP:journals/corr/FinnL16} as well as a premise towards unsupervised learning of visuomotor representations \citep{DBLP:journals/corr/AgrawalNAML16,DBLP:journals/corr/PintoGHPG16}. 
Several formulations have been proposed, under various names, such as world models \citep{DBLP:journals/corr/abs-1803-10122}, action-conditioned video prediction \citep{NIPS2015_5859}, forward models \citep{Miall:1996:FMP:246663.246676}, neural physics \citep{DBLP:journals/corr/ChangUTT16}, neural simulators, etc. A central question is the representation space in which predictions are carried out. 
We identify two main  research threads: 

 (i)  Methods that \textbf{predict the future in a 2D projective space}, such as  future visual frames \citep{DBLP:journals/corr/MathieuCL15,actionconditioned,DBLP:journals/corr/FinnGL16,DBLP:journals/corr/abs-1812-00568},  neural encodings of future frames \citep{DBLP:journals/corr/abs-1803-10122,DBLP:journals/corr/ChiappaRWM17,hafner2019planet,DBLP:journals/corr/AgrawalNAML16,DBLP:journals/corr/PathakAED17},  
 object 2D motion trajectories \citep{Fragkiadaki2015LearningVP,DBLP:journals/corr/BattagliaPLRK16,DBLP:journals/corr/ChangUTT16},  
 2D pixel motion fields 
 \citep{DBLP:journals/corr/FinnL16,DBLP:journals/corr/abs-1710-05268,DBLP:journals/corr/FinnGL16,SE3-Nets}. 
 Models that predict motion and use it to warp pixel colors forward in time need to handle occlusions as well as changes in size and aspect ratio. Though increasing the temporal memory from where to borrow pixel colors helps \cite{lai2020mast,DBLP:journals/corr/abs-1710-05268}, the model needs to learn many complex relationships between pixels to handle such changes over time. 
Recent architectures incorporate object-centric and relational biases in order to generalize across  varying number of objects present in the image, using graph neural networks  \cite{battaglia2018relational,DBLP:journals/corr/abs-1812-10972,ye2019ocmpc,DBLP:journals/corr/BattagliaPLRK16,kipf2018neural}.  
Cross-object interactions are captured 
using edges in an entity graph, where 
messages are iteratively exchanged between the nodes to update each other's embedding  \citep{DBLP:journals/corr/BattagliaPLRK16,pmlr-v80-sanchez-gonzalez18a}. 
Most works that use object factorization biases  either assume the objects' segmentation masks are given or are trivial to obtain from color segmentation \citep{Fragkiadaki2015LearningVP,DBLP:journals/corr/BattagliaPLRK16,DBLP:journals/corr/ChangUTT16}. 
These models work well under a static and fixed camera across training and test conditions, but cannot effectively generalize across camera viewpoints \cite{sudeep} as we empirically validate.

(ii) methods that \textbf{predict the future in a 3D space} of object or particle 3D locations and poses extracted from the RGB images using  human annotations \citep{DBLP:journals/corr/abs-1808-00177} or assumed given   \citep{WuLuKohli17,li2018learning,mrowca2018flexible}. 
 Such explicit 3D state representations are hard in general to obtain from raw RGB input in-the-wild, outside  multiview environments \citep{DBLP:journals/corr/abs-1808-00177, li2018learning}. The work of \cite{particlevisual} attempts to extract the particle 3D locations directly from images, however the encoder proposed is specific to the scene. Different encoders are trained for different scene image to particle mappings. %

 Our work  builds upon  learnable 2D-to-3D convolutional encoders that extract 3D scene representations from images and are trained for self-supervised view prediction, along with tasks of 3D object detection and 3D motion forecasting, relevant for 3D object dynamics learning. Our image-to-3D scene encoders and object detectors generalize across object appearance and number of objects, and do not assume any ground-truth information about the object or particle locations \cite{particlegt,particlegt2,particledense} or image segmentation \cite{wu2017learning} during test time.

\section{Object-Factorized Environment Simulators (3D-OES)}
\vspace{-0.1in}
\label{sec:model}
\begin{figure}[t]
	\centering
		\centering
		\includegraphics[width=0.92\textwidth]{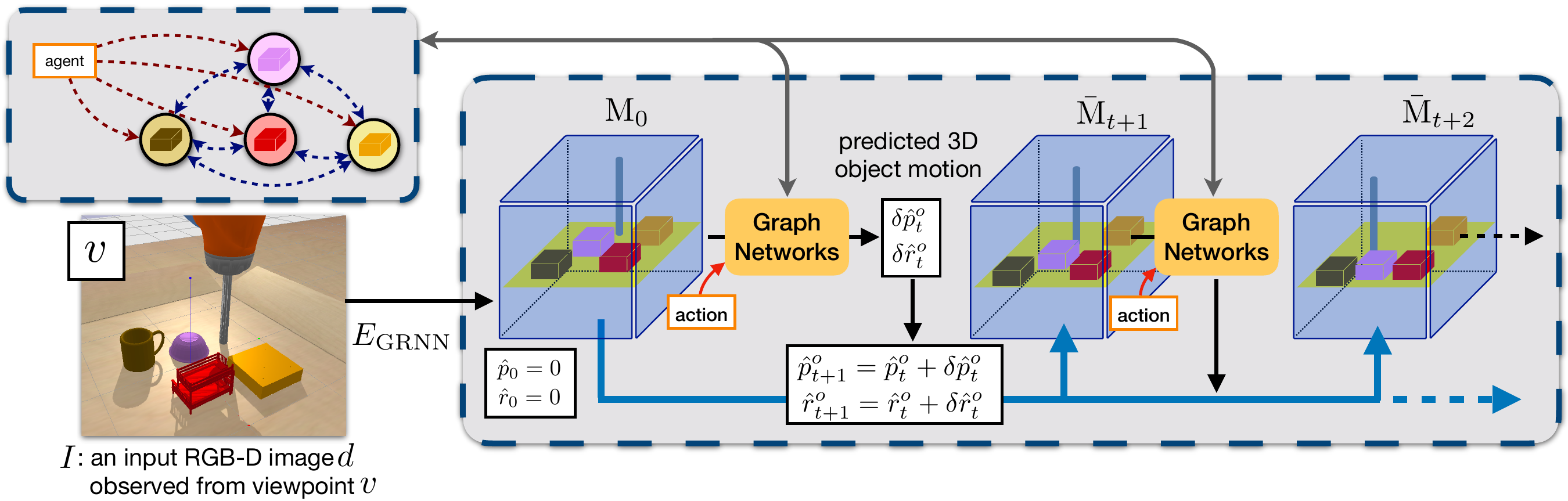}
		\caption{
		\textbf{\ES} predict 3D object motion under agent-object and object-object interactions, using a graph neural network over 3D feature maps of detected objects. Node features capture the appearance of an object node and its immediate context, and  edge features capture relative 3D locations between two nodes, so  the model is translational invariant. After message passing between nodes, the node and edge features are decoded to future 3D rotations and translations for each object.
		}
			\label{fig:model}\vspace{-12pt}
\end{figure}

The architecture of \ES~ is depicted in Figure \ref{fig:model}.  
At each time step, our model takes as input a single or a set of  RGB-D images of the scene along with the corresponding camera views to capture them, and encodes these inputs into a 3D scene feature representation by neurally mapping image and depth maps to 3D feature grids (Section \ref{sec:enc}). Then, it detects 3D object boxes  in the inferred 3D  scene representation, and crops the scene representation to obtain a set of  object-centered 3D feature maps.  A graph neural network over the object nodes will take as inputs object appearances and the agent actions and predict the future 3D rotation and translation   for each object (Section \ref{sec:forecast}). We will assume for now rigid objects, and we  discuss in Section \ref{sec:conclusion} how to extend our framework to  deformable and articulated  objects. 
Our model generates future  scenes by
warping  object-centric 3D feature maps with the predicted cumulative  3D object motion (Section \ref{sec:forecast}).  These synthesized future 3D scene feature maps, though not directly interpretable, can be decoded to RGB images from any desired camera viewpoints via a neural renderer to aid interpretability.
We use long-term simulations of \ES\  to generate  action plans for pushing objects to desired locations in cluttered environments using model predictive control (Section \ref{sec:control}).  
We apply our model to learn dynamics of objects  pushed around on a table surface and objects falling on top of others.  
At training time, we  assume access  to  3D object bounding boxes to train our 3D object detector. 

\vspace{-0.1in}
\subsection{Differentiable 2D-to-3D lifting with  Geometry-Aware Recurrent Networks (GRNNs)}
\label{sec:enc}
Geometry-Aware Recurrent Networks (GRNNs) introduced in \cite{commonsense,adam3d} are network architectures equipped with a differentiable unprojection (2D-to-3D) module and a 3D scene neural map as their   bottleneck.  They can be trained end-to-end for a downstream task, such as supervised 3D object detection or self-supervised view prediction. 
   We will denote the 3D scene feature   as $\M \in \mathbb{R}^{w \times h \times d \times c}$ where $w,h,d,c$ denote width, height, depth and number of channels, respectively. 
 Every $(x, y, z)$ grid location in the 3D feature map $\M$ holds a $c$-dimensional feature vector  that  describes the semantic and geometric properties of a corresponding physical location in the 3D world scene. 
 Given an input video, GRNNs estimate the relative camera poses between frames, and
 transform the inferred 3D features map $\M_t$ to a world coordinate frame to cancel the camera egomotion, before accumulating it with  3D feature maps across time steps. 
 In this way,    information from 2D pixels that correspond to the same 3D physical point end up nearby in the 3D neural map. We use such cross-view registration in case we have access to concurrent multiple camera views for the first timestep of our simulations.  
     Upon training, GRNNs  map  RGB-D images or a single RGB-D image to a  \textit{complete} 3D feature map of the scene they depict, i.e., the model learns to \textit{imagine} the missing or occluded information from the input view. We denote this 2D-to-3D mapping as $\M=\Enc(I_1, ..., I_t)$, where $\M \in \mathbb{R}^{w \times h \times d \times c} $ and $I_t = \{d_t, v_t\}$ denotes the RGB-D image $d_t$ and the corresponding camera pose $v_t$ at time step $t.$ Note that the input can be a single RGB-D view, in which case $\M=\Enc(I)$.   For further details on GRNNs, please refer to \cite{commonsense,adam3d}.

\noindent   \textbf{View prediction for visualizing  latent 3D neural simulations}
  We   train GRNNs end-to-end for RGB view regression in videos of static scenes and moving cameras as proposed in \cite{commonsense}, by neurally projecting the 3D scene feature maps and mapping them to 2D images. Our decoder involves a differentiable 3D-to-2D projection module that projects the 3D scene feature representation after orienting it to the query camera viewpoint. The projected features are then decoded into images through a learned decoder. In this way, the trained projection and decoding module can be used to interpret and  visualize the 3D latent feature space with view-specific 2D images, given any desired camera viewpoint.

\noindent \textbf{3D object detection }
  Our model uses a 3D object detector to map the 3D scene neural map $\M$ to a variable number of object axis-aligned 3D boxes and corresponding 3D segmentation masks, i.e., binary 3D voxel occupancies:  $\mathcal{O}=\Det(\M), \mathcal{O} =\{\hat{b}^o=(p^o_x,p^o_y,p^o_z,w^o,h^o,d^o)\in \mathbb{R}^6, m^o \in \{0,1\}^{w^o \times h^o \times d^o}, o=1 \cdots |\mathcal{O}| \}$, where $p^o_x,p^o_y,p^o_z$ stands for the 3D box centroid and $w^o,h^o,d^o$ stands for 3d box size. Its architecture is similar to Mask R-CNN \cite{DBLP:journals/corr/HeGDG17} but uses 3D input and output instead of 2D. Given an object 3D centroid $p^o_x,p^o_y,p^o_z$, we crop the 3D scene feature map $\M$ using a corresponding fixed-size axis-aligned 3D bounding box  to obtain corresponding object-centric feature maps $\M^o, o=1 \cdots |\mathcal{O}|$ for all objects in the scene.

\vspace{-0.1in}
\subsection{3D Object Graph Neural Networks for Motion Forecasting} \label{sec:forecast}
\vspace{-2mm}
Objects are  the recipient of forces exercised by active agents; meanwhile, objects themselves  carry momentum and cause other objects to move. 
How can we model cross-object dynamic relationships in a way that generalizes with varying number of objects and arbitrary chains of interactions? 

We consider a graph interaction network \cite{DBLP:journals/corr/BattagliaPLRK16} over the graph comprised of the detected objects and the agent's end-effector. Inputs to the network are the object-centric feature maps, one per object node, the objects' velocities, the agent's action represented as a 3D translation, as well as edge features, which incorporate the relative 3D displacements between the nodes.  The outputs of the network are the  3D translations $\delta \hat{p}$ and 3D relative rotations $\delta \hat{r}$ of the object nodes at the next time step.  
During message passing in the constructed graph, edge and node features are encoded and concatenated, and messages from neighboring nodes are aggregated via summation. 
Our graph network is trained supervised to minimize a standard regression loss for the next time step.

\vspace{1mm}\textbf{Forward unrolling with object appearance permanence} To predict long term results of actions, as well as results of action sequences, the model needs to be unrolled forward in time as commonly done in related works \cite{battaglia2018relational,DBLP:journals/corr/abs-1812-10972,ye2019ocmpc,DBLP:journals/corr/BattagliaPLRK16,kipf2018neural}. Different from previous works though, 
\ES\ can synthesize 3D neural scenes of future timesteps  by  warping (translating and rotating)  object feature maps obtained from the first timestep---as opposed to the ones obtained from the predicted scene of the previous timestep---according to cumulative 3D motion predictions. 
Specifically, given predicted 3D object motions ($\delta \hat{p}_t$, $\delta \hat{r}_t$) at an unrolling step $t$, we estimate the \textbf{cumulative} 3D rotation and translation of the object with respect to the first timestep:
\begin{align}
&\hat{p}_t=\hat{p}_{t-1}+\delta \hat{p}_t, \ \   
\hat{\rot}_t=\hat{r}_{t-1}+\delta \hat{\rot}_t , \ \   t=1 \cdots T, \ \  \hat{p}_0=0 \ \ \hat{\rot}_0 = 0.
\end{align}
where $T$ denotes the number of unrolling steps thus far. 
Then, given 3D object segmentation masks $m^o$ and object-centric 3D feature maps $\M^o$ obtained by the 3D object detector from the input RGB-D image, we rotate and translate the object masks and 3D  feature maps  using the cumulative 3D rotation $\hat{\rot}_t$ and 3D translation $\hat{p}_t$  using  3D spatial transformers. We  synthesize a new 3D scene feature map $\bar{\M}_{t}$ by placing each transformed object-centric 3D feature map at its predicted 3D location: 
     $  \bar{\M}_{t} = \sum_{o=1}^{|\mathcal{O}|} \mathrm{Draw}(\mathrm{Rotate}(m^{o}, \hat{\rot}^o_t) \odot \mathrm{Rotate}(\M^{o}, \hat{\rot}^o_t), \hat{p}^o_{t}), $
where superscript $o$ denotes the object identity, $\mathrm{Rotate}(\cdot, r)$ denotes 3D rotation by  angle $\rot$,  $\odot$ denotes voxel-wise multiplication,  and  $\mathrm{Draw}(\M, p)$ denotes adding a feature tensor $\M$  at a 3D location $p$. 
This synthesized  scene map is used for neural rendering to help interpret the predicted scene at $t$. To obtain the inputs for our graph neural network at the next time step, we can potentially 
crop the synthesized 3D scene map $\bar{\M}_{t}$ at the predicted 3D location. However, we find that directly using object features obtained in the first time step and including accumulative relative object pose as part of the object state works better in practice.

Our graph neural motion forecaster is trained through forward unrolling. Error of each time step is back-propagated through time.
 More implementation details are included in Appendix Section C.1.

\vspace{-0.1in}
\subsection{Model Predictive Control with \ES}
\label{sec:control}
\vspace{-2mm}
Action-conditioned  dynamics models, such as \ES, simulate the  results of an agent's actions and permit successful control in zero-shot setups: achieving a specific  goal in a novel scene without previous practice. 
We apply our model for pushing objects to desired locations in cluttered environments with model predictive control. Given an input RGB-D image $I$ that contains  multiple objects, a goal configuration is given in terms of the desired 3D location of an object $\x_{goal}^o$. 
\ES\ infer the scene 3D feature map $\M=\Enc(I)$ and detects the objects present in the  scene. We then unroll the model  forward in time using randomly sampled action sequences, as described in Section \ref{sec:forecast}.  
We evaluate each action sequence based on the Euclidean distance from the goal to the predicted location $\hat{x}^o_T$ (after $T$ time steps) for the designated object. We execute the first action of the best action sequence and repeat  \citep{tassa2008receding}. 
Our model combines 3D perception and planning using learned object dynamics in the inferred 3D scene feature map. 
While most previous works choose bird's eye viewpoints to minimize cross-object or robot-object occlusions \cite{fang2019dynamics}, our control framework \textbf{can use any camera viewpoint}, thanks to its ability  to map input 2.5D images to complete, viewpoint-invariant  3D scene feature maps. We empirically validate this claim in our experimental section.

\section{Experiments}\label{sec:experiments}
We evaluate our model on its prediction accuracy for single- and multi-step object motion forecasting under multi-object interactions, as well as on its performance in model predictive control for pushing objects to desired locations on a table surface in the presence of obstacles. 
 We ablate generalization of our model under varying camera viewpoints and varying number of object and varying object appearance.  
 Our model is trained to predict 3D object motion during robot \textbf{pushing} and \textbf{falling} in the Bullet Physics Simulator.
For \textbf{pushing}, we have objects pushed by a Kuka robotic arm and record RGB-D video streams from multiple viewpoints. We create scenes using 31 different 3D object meshes, including 11 objects from the MIT Push dataset \citep{mitpush} and 20 objects randomly selected from \textit{camera}, \textit{mug}, \textit{bowl}, and \textit{bed} object categories  of the ShapeNet dataset \citep{shapenet2015}.  At training time, each scene contains at most \textit{two} objects. We test with varying number of objects. 
For \textbf{falling}, we use 3D meshes of the objects introduced in \cite{DBLP:journals/corr/abs-1812-10972}, including a variety of shapes. We randomly select 1-3 objects and randomly place them on a table surface, and let one object fall  from  a height. 
We train our model with three camera views, and use either three or one randomly selected views as input during test time.

We compare \ES\  against a set of baselines designed to cover representative models in the object dynamics literature:
\textbf{(1)} \Bone, a model that mimics Interaction Networks  \citep{DBLP:journals/corr/BattagliaPLRK16} and \citep{DBLP:journals/corr/abs-1808-00177, WuLuKohli17}. It is a graph neural network in which object features are the 3D object centroid locations and their velocities, and edge features are their relative 3D locations. 
    \textbf{(2)}  \Btwo, a model using graph neural network over 3D object centroid locations and object-centric 2D image CNN feature embeddings, similar to \cite{ye2019ocmpc}. The model further combines camera pose information with the node features. 
\textbf{(3)} 
    Visual Foresight (\textit{VF}) \cite{ebert2018visual}, a model that uses the current frame and the action of the agent to predict future 2D frames by ``moving pixels" based on predicted 2D pixel flow fields.
\textbf{(4)} 
    \Pla \ \cite{hafner2019planet}, a model that learns a scene-level embedding by predicting future frames and the reward given the current frame. 

We compare our model against baselines \Bone \ and \Btwo \  on both motion forecasting and model predictive control. Since \VF \ and \Pla \ forecast 2D pixel motion and do not predict explicit 3D object motion, we  compare against them on the pushing task with model predictive control. For further details on data collection, train-test data split, and implementation of the baselines, please refer to Appendix (Section B and C.2). 

\begin{table}[h!]
\centering
\caption{\textbf{3D object motion prediction test error during object pushing  in scenes with two objects} for 1,3, and 5 timestep prediction horizon. 
}
\begin{tabular}{c|l|cccc}
            \multicolumn{1}{l|}{Experiment Setting}          & 
            \multicolumn{1}{l|}{Model}          & 
            \multicolumn{1}{l}{}          & 
            \multicolumn{1}{l}{T=1} & \multicolumn{1}{l}{T=3} &  \multicolumn{1}{l}{T=5}
            \\ \hline
3 views (random, novel)  &\Bone~  

\citep{DBLP:journals/corr/BattagliaPLRK16}& translation(mm) 
                 & $4.6$ & $32.1$  & $66.3$  \\
       + gt-bbox &      & rotation(degree) 
                 & $2.8$ & $16.7$ & $26.4$\\
                 &\Btwo  ~\citep{ye2019ocmpc}&translation(mm)  
                 & $6.0$& $39.3$& $69.7$\\
                 &      & rotation(degree) 
                 &$3.4$ & $29.8$  & $30.7$  \\
                 &Ours  &translation(mm)
                 & ${\bf 3.6}$ & $ {\bf 22.5}$ & ${\bf 43.4}$ \\
                 &      & rotation(degree) 
                & ${\bf 2.5}$ & ${\bf 12.0}$ & ${\bf 20.6}$ \\\hline
1 view (random, novel) &\Btwo ~\citep{ye2019ocmpc} &translation(mm)  
                 & $6.0$& $39.3$& $69.7$\\
     + gt-bbox           &      & rotation(degree) 
                 &$3.4$ & $29.8$  & $30.7$  \\
                 &Ours  &translation(mm)
                 &  ${\bf 4.1}$ & ${\bf 23.6}$ & ${\bf 43.8}$ \\
                 &      & rotation(degree) 
                & ${\bf 3.1}$ & ${\bf 12.2}$ & ${\bf 20.3}$ \\\hline
1 view (random, novel) &\Bone ~  
\citep{DBLP:journals/corr/BattagliaPLRK16}& translation(mm) 
                 & 6.7 & 35.4 & 68.2  \\
                 + predicted-bbox   &      & rotation(degree) 
                 & 3.0 & 20.1 & 30.32 \\
                 &\Btwo ~\citep{ye2019ocmpc} &translation(mm)  
                 & $6.6$& $43.1$& $71.2$\\
                 &      & rotation(degree) 
                 &$3.6$ & $31.8$  & $32.4$  \\
                 &Ours  &translation(mm)
                 & ${\bf 4.3}$& ${\bf 25.2}$& ${\bf 47.0}$\\
                 &      & rotation(degree) 
                 & ${\bf 2.7}$ & ${\bf 12.1}$ & ${\bf 19.7}$  \\\hline
1 view (\textbf{fixed, same as train}) 
                 &\Btwo ~\citep{ye2019ocmpc}  &translation(mm)  
                 & $5.1$& $29.6$ & $54.5$\\
+ predicted-bbox 
                 &      & rotation(degree) 
                 & $2.6$ & $11.0$ & $16.9$ 
               \\\hline
\end{tabular}

\label{tab:2obj}
\end{table}

\begin{table}[h!]
\caption{\textbf{3D object motion prediction test error during object falling  in scenes with three to four objects} for 1,3, and 5 timestep prediction horizon.
}
\centering
\begin{tabular}{c|l|cccc}
            \multicolumn{1}{l|}{Experiment Setting}          & 
            \multicolumn{1}{l|}{Model}          & 
            \multicolumn{1}{l}{}          & 
            \multicolumn{1}{l}{T=1} & \multicolumn{1}{l}{T=3} &  \multicolumn{1}{l}{T=5}
            \\ \hline
1views (random, novel)  &\Bone ~  

\citep{DBLP:journals/corr/BattagliaPLRK16} & translation(mm) 
                 & $5.2 $ & ${\bf 11.7}$  & $278.6$  \\
       + predicted-bbox &      & rotation(degree) 
                 & $ {\bf 5.7} $ & $ {\bf 10.4} $ & $ 43.28 $\\
                 &\Btwo ~\citep{ye2019ocmpc} &translation(mm)  
                 & $8.4$& $17.0$& $620.2$\\
                 &      & rotation(degree) 
                 &$9.2$ & $16.6$  & $117.9$  \\
                 &Ours  &translation(mm)
                 & ${\bf 5.0}$ & $ 13.1$ & ${\bf 16.4}$ \\
                 &      & rotation(degree) 
                & $ 6.1$ & $ 12.6 $ & ${\bf 18.7}$ \\\hline
\end{tabular}

\label{tab:2obj_fall}
\end{table}

\subsection{Action-Conditioned 3D Object Motion Forecasting}
We evaluate the performance of our model and the baselines in single- and multi-step 3D motion forecasting for \textbf{pushing} and \textbf{falling} on novel objects in Tables \ref{tab:2obj} and \ref{tab:2obj_fall} in terms of translation and rotation error. 
We evaluate the following ablations: i) using 1 or 3 camera views at the first time step, ii) using goundtruth 3D object boxes (gt-bbox) or 3D boxes predicted by our 3D detector, iii) varying camera viewpoints (random) versus keeping a single fixed camera viewpoint at train and test time. 
Our model outperforms the  baselines both in translation and rotation prediction accuracy. When tested with object boxes predicted by the 3D object detector (Section \ref{sec:enc}) as opposed to ground-truth 3D boxes, our model is the least affected. \Btwo \  performs on par with or even worse than \Bone, indicating that it does not gain from having access to additional appearance information.  We hypothesize this is due to the way appearance and camera pose information are integrated in this baseline: the model simply treats camera pose information as additional input, as opposed to our model, which leverages geometry-aware representations that retain the geometric structure of the scene.

\begin{figure*}[t!]
	\centering
		\centering
		\includegraphics[width=1.0\textwidth]{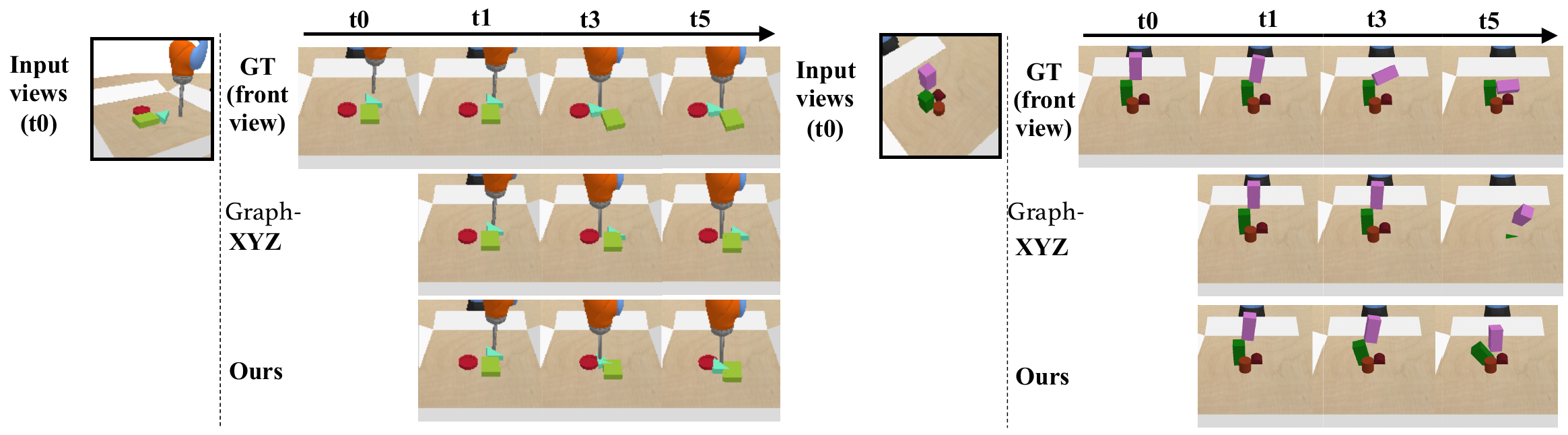}
		\caption{{\bf Forward unrolling of our dynamics model and the \Bone\  baseline.} Left: pushing. Right: falling.
		In the top row, we show (randomly sampled) camera  views that we use as input to our model. 
		The second row shows the ground-truth motion of the object from the front view. 
		Rows 3, 4 show the predicted object motion from our model and the \Bone \  baseline from the same front camera viewpoint. Our model better matches the ground-truth object motion than the \Bone \  baseline. The latter does not capture object appearance in any way. 
		}
			\label{fig:rollout}
\end{figure*}

\textbf{Multi-step forward unrolling} 
The \Bone \  baseline can be easily  unrolled  forward in time without  much error accumulation since it does not use any appearance features. Still, as seen in Tables \ref{tab:2obj} and \ref{tab:2obj_fall}, our model outperforms it.
\Bone \ is oblivious to the appearance of the object and thus cannot effectively adapt its predictions to different object shapes.

\textbf{Varying number of camera views} 
 Our model  accepts a variable number of  views as input, and improves when more views  available; yet, it can accurately predict future motion even from a single RGB-D view. 
 The prediction error of our single view model is only slightly higher than the model using three random views as input.  As shown in Table \ref{tab:2obj}, the \Btwo \  baseline performs the worst and does not improve with more views are available. We believe this is due to the geometry-unaware way of combining multiview information by concatenation, though the model does have access to camera poses of the input images. 
 
 \textbf{Varying camera viewpoint versus fixed camera viewpoint} 
 We show in Table \ref{tab:2obj} (last 2 rows) that \Btwo \  can achieve much better  performance when trained and tested on a single fixed camera viewpoint. This is a setting widely used in the recently popular learning-based visual-motor control literature \cite{DBLP:journals/corr/FinnL16, pathakICLR18zeroshot, ye2019objectcentric, ebert2018visual}, which restricts the corresponding models to  work only under carefully controlled environments with a fixed camera viewpoint, while ours performs competitively to these model but also handles arbitrary camera viewpoints.


\textbf{Visualization of the 3D motion predictions}
In Figure \ref{fig:rollout}, we  show qualitative long term motion prediction results produced by unrolling our model forward in time (more are shown in the supplementary video).
Our model generalizes to novel objects and scenes with varying number of objects, though trained only on 2 object scenes. 

\begin{figure*}[h!]
	\centering
		\centering
		\includegraphics[width=1.0\textwidth]{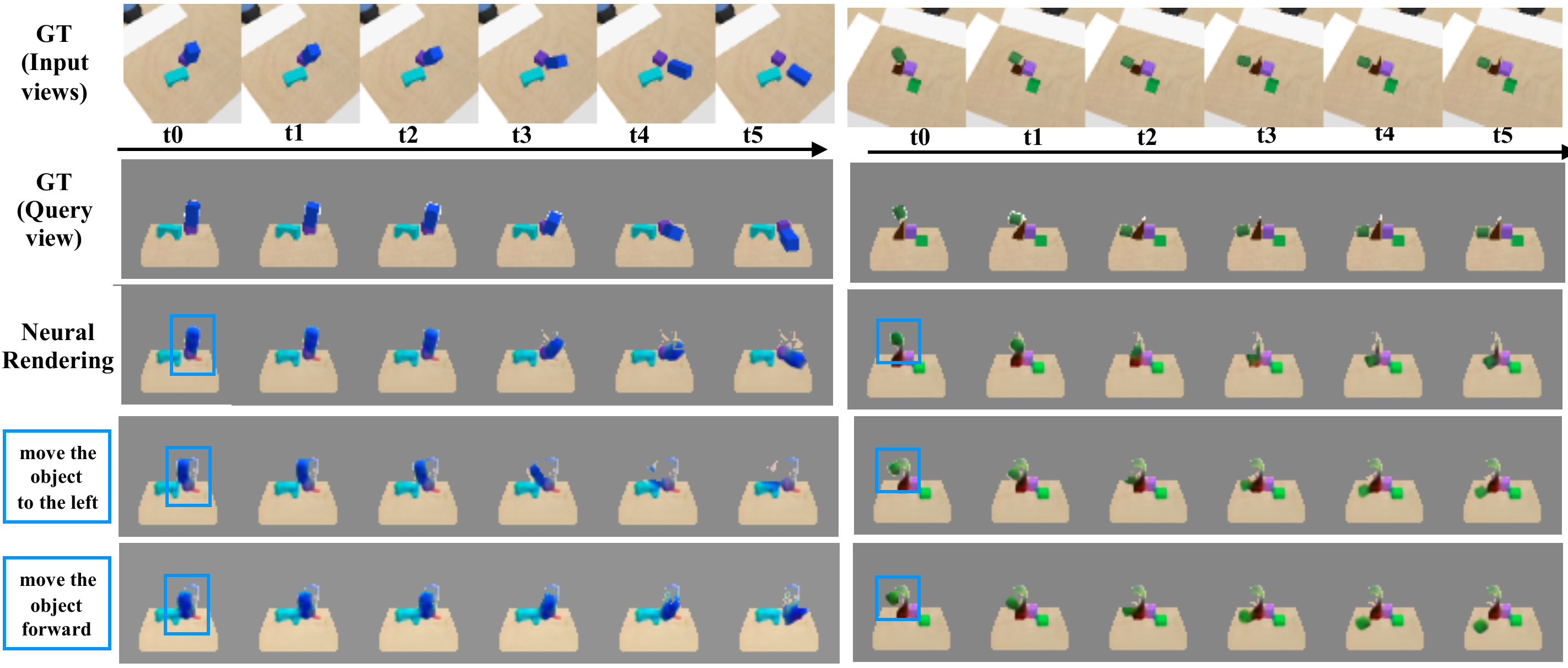}
		\caption{{\bf Neurally rendered simulation videos of counterfactual experiments.} The first row shows the ground truth simulation video from the dataset. Only the first frame in this video is used as input to our model to produce the predicted simulations. The second row shows the ground truth simulation from a query view. The third row shows the future prediction from our model given the input image. The following rows show the simulation after manipulating an objects (in the blue box) according the instruction on the left most column.
		}
			\label{fig:conterfectual2_2}
\end{figure*}

\textbf{Neural rendering and counterfactual simulations}
3D-OES not only can simulate the future state of the scene, it also provides us a way to interpret the latent 3D representation and a space to run counterfactual experiements. We visualize the latent 3D feature map by neurally projecting it from a camera viewpoint to an image through a learned neural decoder, and show the resulting images in Figure \ref{fig:conterfectual2_2}. We also show that our 3D representation allows us to alter the observed scene and run conterfactual simulations in multiple ways. More results are provided in Appendix Section A.

\subsection{Pushing with Model Predictive Control (MPC)}
We test \ES \  on pushing objects to desired locations using MPC and report the results in Table \ref{tab:control}. For our model and \Btwo, we use a single randomly sampled input view. For \VF \  and \Pla, we use a fixed top-down view for both training and testing as we found they only work reasonably well with a fixed viewpoint. 
Details of the experiment settings are included in Appendix (Section C.3).
Our model outperforms all baselines by a large margin. 
We include videos of pushing object to desired locations in the presence of multiple obstacles in the supplementary file. 

\begin{table}[t!]
\caption{Success rate for pushing objects to target locations. }
\centering
\begin{tabular}{c c c c c|c}
            \multicolumn{1}{l}{\Bone ~  
\citep{DBLP:journals/corr/BattagliaPLRK16}}        &  \multicolumn{1}{l}{\Btwo ~\citep{ye2019ocmpc}} & \multicolumn{1}{l}{\VF \cite{ebert2018visual} } & \multicolumn{1}{l}{\Pla \cite{hafner2019planet} }  &
            \multicolumn{1}{l|}{Ours} &\multicolumn{1}{l}{Ours-Real}  \\ \hline
0.76 & 0.70  & 0.32 & 0.16& ${\bf 0.86}$ & 0.78 \\
\end{tabular}
\label{tab:control}
\end{table}

\begin{figure*}[t!]
	\centering
		\centering
		\includegraphics[width=1.0\textwidth]{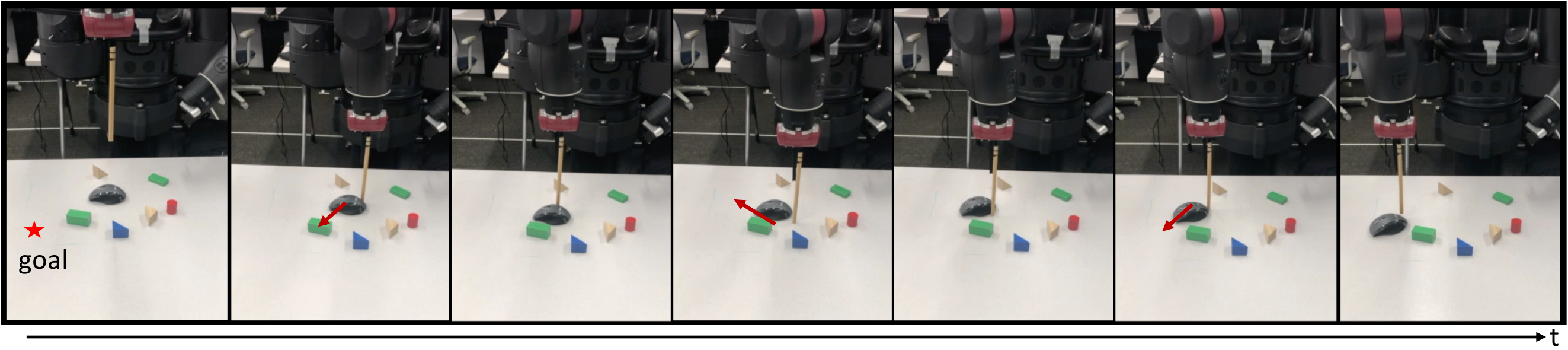}
		\caption{{\bf Collision-free pushing on a real-world setup.} 
		The task is to push a mouse to a target location without colliding into any obstacles. Our robot can successfully complete the task with 3 push attempts.
		}
		\label{fig:robot}
	\vspace{-3mm}
\end{figure*}

\textbf{Sim-to-Real Transfer}
We train our  model solely in simulation and test  it  on object pushing control tasks on a real  Baxter platform equipped with a rod-shaped end-effector, similar to the setting in the Bullet simulation (Figure \ref{fig:robot}). We attached a Intel RealSense D435 RGB-D camera to the robot's left hand, and use only one RGB-D view as input for this experiment. The pose of the camera is different from those seen during  training. Please refer to Appendix (Section C.3) for details of our real-world setup, objects selection, and 3D detector training. 
We report the success rate of real-world pushing in Table \ref{tab:control} (Ours-Real). Our model achieves similar success rates for pushing  in simulation and in the real world. Since geometry information is shared by simulation and the real world by a large extent, and our model combines the viewpoint-invariant property of the geometry-aware representation and an object-factorized structure, it presents good sim-to-real transferrability. 
 In Figure \ref{fig:robot}, we qualitatively show pushing objects along collision-free trajectories in complex scenes in the real-world setup. More examples are included in the supplementary video.

\vspace{-1mm}
\section{Conclusion} \label{sec:conclusion}
\vspace{-1mm}
We have presented \ES\ , dynamics models that  predict 3D object  motion  in a 3D latent visual feature space inferred from 2.5D video streams. 
We empirically showed our model can generalize across camera viewpoints and varying number of objects better than existing 2D dynamics models or dynamics models over 3D object centroids. To the best of our knowledge, this is the first model that can predict 3D object dynamics directly from RGB-D videos and generalize across scene variations.  

Our model currently has  three main limitations:
\textbf{(i)} It requires ground-truth 3D object locations and orientations at training time.
Automatically inferring those with 3D tracking 
 would permit our  model to be trained in a  self-supervised manner.
\textbf{(ii)} It assumes rigid object interactions.
Learning dynamics of soft bodies and fluids would require  forecasting   dense 3D motion fields, or considering sub-object (particle) graphs. 
\textbf{(iii)} It is deterministic. Handling stochasticity
via  stochastic models or objectives would permit to learn more complex and multimodal object motions.


\textbf{Acknowledgements} This paper is based upon work supported by  Sony AI, ARL W911NF1820218, DARPA Common Sense program,  NSF Grant No. IIS-1849287, and an NSF CAREER award. Fish Tung is supported by Yahoo InMind fellowship and Siemens FutureMaker Fellowship.

\newpage

{
\setlength{\bibsep}{2pt}
\small
\bibliography{8_refs}
}

\appendix
\section*{Appendix}

\section{Additional experimental results}

\subsection{Neurally rendered physics simulations from multiple views}

\begin{figure*}[h!]
	\centering
		\centering
		\includegraphics[width=1.0\textwidth]{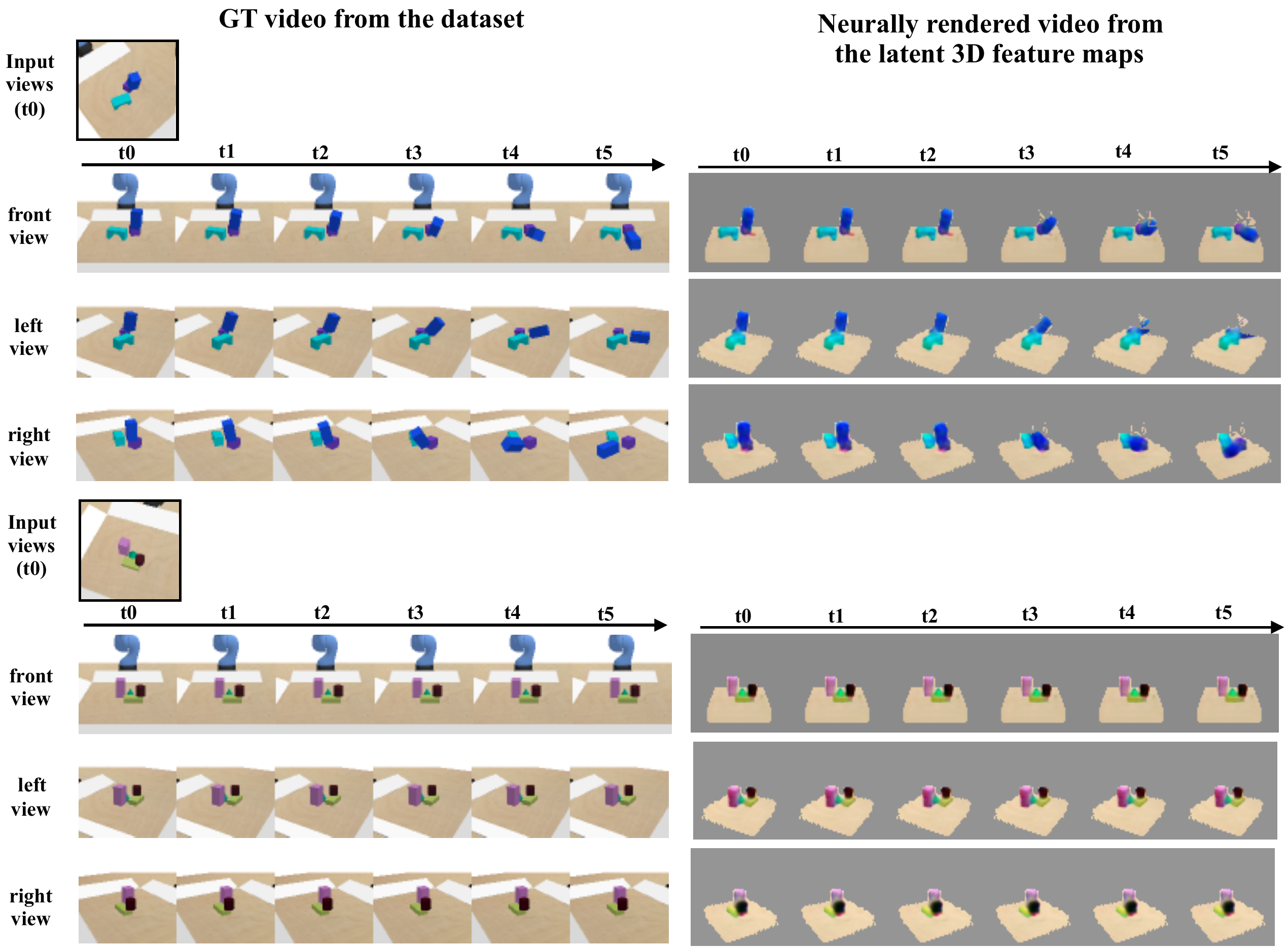}
		\includegraphics[width=1.0\textwidth]{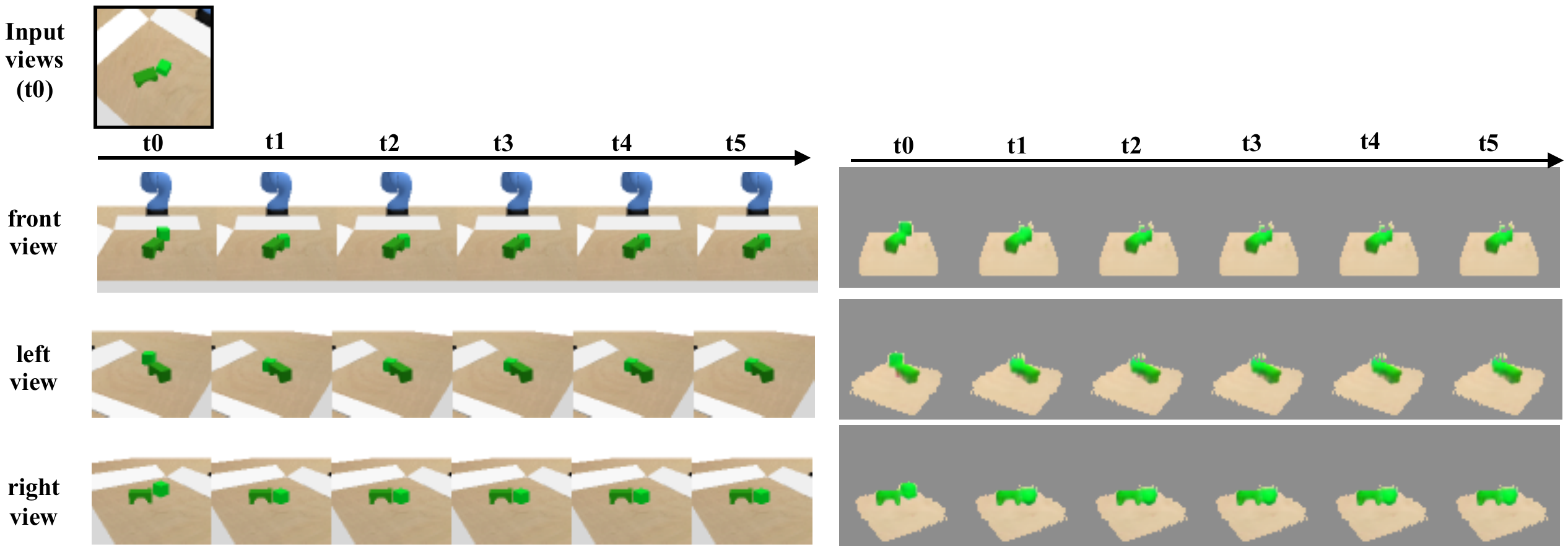}
		\caption{{\bf Neurally rendered simulation videos from three different views} Left: groundtruth simulation videos from the dataset. The simulation is generated by the Bullet Physics Simulation. Right: neurally rendered simulation video from the proposed model. Our model forcasts the future latent feature by explicitly warping the latent 3D feature maps, and we pass these warped latent 3D feature maps through the learned 3D-to-2D image decoder to decode them into human interpretable images. We can render the images from any arbitrary views and the images are consistent across views.
		}
			\label{fig:neural_render2}
\end{figure*}

We show in Figure \ref{fig:neural_render2} rendered physics simulation videos using the proposed model. The latent 3D feature map of the proposed model is interpretable in the sense that we can render human-interpretable RGB images 
from the feature map using the learned neural image decoder. More importantly, we can render such simulation videos from any arbitrary view, and the videos captured from different views are consistent with each other.

\subsection{Additional counterfactual experiments}

\begin{figure*}[h!]
	\centering
		\centering
		\includegraphics[width=1.0\textwidth]{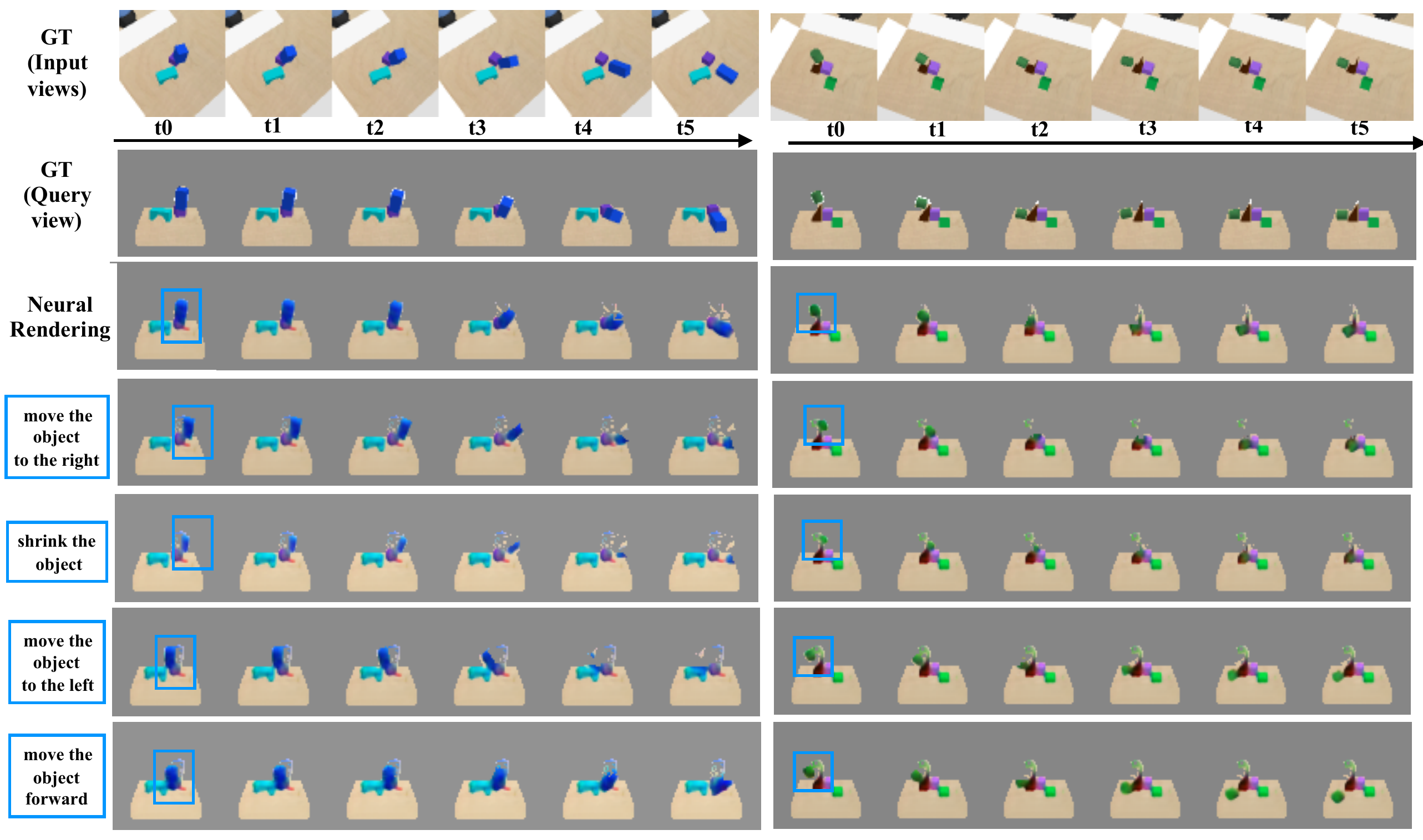}
		\caption{{\bf Neurally rendered simulation videos of counterfactual experiments.} The first row shows the ground truth simulation video from the dataset. Only the first frame in this video is used as input to our model to produce the predicted simulations. The second row shows the ground truth simulation from a query view. Note that our model can render images from any arbitrary view. We choose this particular view for better visualization. The third row shows the future prediction from our model given the input image. The following rows show the simulation after manipulating an objects (in the blue box) according the instruction on the left most column.
		}
			\label{fig:conterfectual2}
\end{figure*}

In Figure \ref{fig:conterfectual2}, we show more results of conducting counterfactual experiments using the learned neural simulator. We can move objects to arbitrary position, and change their size by moving their features explicitly in the latent 3D feature space. Although the model has never been trained on this task, it can generate reasonable simulation results after such manipulations on the objects.

\section{Data collection details}
Here we describe details of the data used in Section 4.
\newline

\noindent\textbf{Pushing} Our training  data contains RGB-D video streams where the robot pushes objects which in turn can collide and push other objects on the table.
We create scenes using 31 different 3D object meshes, including 11 objects from the MIT Push dataset \citep{mitpush} and 20 objects selected from four categories (\textit{camera}, \textit{mug}, \textit{bowl} and \textit{bed}) in the ShapeNet Dataset \citep{shapenet2015}. We split our dataset so that 24 objects are used during training. At test time, we evaluate the prediction error on the remaining 7  objects. 
At training time each scene contains at most two (potentially interacting) objects. At test time, we vary the number of objects from one up to five. We randomize the textures of the objects during training to improve transferability to the real world \cite{DBLP:journals/corr/TobinFRSZA17}. We consider a simulated Kuka robotic arm equipped with a single rod (as shown in Figure 3 of the main paper. The objects can move  on a planar table surface of size $0.6m \times 0.6 m$ when pushed by the arm, or by other objects. We collect training interaction trajectories by instantiating the gripper nearby a (known) 3D object  segmentation mask. We sample random pushing action sequences with length of 5 timesteps, where each action is a horizontal displacement of the robot's end-effector ranging from $3cm$ to $6cm$, and each timestep is defined to be 200ms. We record objects displacement 1 sec after the push. We place cameras at 27 nominal different views including 9 different azimuth angels ranging from the left side of the agent to the right side of the agent combining with 3 different elevation angles from 20, 40, 60 degrees. All cameras are looking at the 0.1m above the center of the table, and are 1 meter away from the look-at point. At each timestep, all cameras are purturbed randomly around their nominal viewpoints, and we record all 27 views. At training time,  our model consumes  three randomly selected concurrent camera viewpoints as input.  At test time, we use the 3D object detector to predict the 3D object segmentation mask, and our model is tested  with either three or a  single view as input, all randomly selected.  All images are $128 \times 128$. There are 5000 pushing trajectories in the training data, and 200 pushing trajectories in the test data.
\newline

\noindent \textbf{Falling}
We use the 3D meshes of the block objects introduced in \cite{DBLP:journals/corr/abs-1812-10972}, which includes cones, cylinders, rectangles, tetrahedrons, and traingles with a variety of shapes. We randomly select 1-3 objects and initialize their position by placing them on the table surface, and let one object falls freely from the air. One timestep is defined to be 40ms. All other settings are identical to the settings for pushing.

\section{Experimental details}
\subsection{Implementation Details of the 3D Object Graph Neural Networks}
We use ground-truth 3D bounding boxes for cropping the scene feature map $\M$ at training time, and 3D predicted boxes provided by our 3D detector at test time. The loss function is the summation of the L2 distance between the predicted and GT translation, and the L2 distance between the predicted and GT quaternions. The inputs to the graph networks are the cropped 3D feature maps of each objects with the size of $16 \time 16 \times 16.$ We first transforms the object-center 3D feature map into a feature vector with three 3D-conv layers followed by an average pooling layer and two FC layers of size 32 with leaky-relu. The vectorized object features are then concatenated with the position and orientation of the objects as inputs to a standard graph network. In the graph network, both the node and edge encoders are 4-layer MLPs with layer size of 32 and leaky-relu activation. Our model is initialized with Xavier initialization and trained using the Adam optimizer for 90K steps. We train two separate models, one for pushing and one for falling.

For the 2D-to-3D image encoding using GRNNs \cite{commonsense}, 
we follows the exact neural architecture as in \cite{commonsense}, which takes as input RGB-D images and outputs 3D feature map $\M_t$ of size $64\times 64\times 64\times32$. Our detector also follows the architecture design of  \cite{commonsense}, which extends the 2D faster RCNN architecture to predict 3D bounding boxes from 3D features maps, as opposed to 2D boxes from 2D feature maps.  The detector takes the output 3D feature map from the 2D-to-3D lifting as input to predict object bounding boxes. 
The detector consists of one down-sampling layer and three 3D residual blocks, each having 32 channels. We use 1 anchor box at each grid location in the 3D feature map with a size of 0.12 meters. The detector predicts an objectness score for each anchor box and selects boxes that exceeds a threshold. We set the threshold to be 0.9. We train the object detector with all the frames in the training data.


\subsection{Implementation Details for Baselines}
Here we describe the baselines discussed in Section 4 in detail.
\begin{enumerate}
    \item  \Bone, a model that uses the 3D object centroid (X, Y, Z)  as object state, and incorporate cross-object interactions for forecasting 3D translation using graph convolutions  over a object graph, similar to  \citep{DBLP:journals/corr/abs-1808-00177, WuLuKohli17}. Since the canonical pose of an object is undefined, object orientation is not included in the object state. 
    This model neglects object shape and appearance. 
    The graph networks used in all baselines follow the exact design as the one we use in our model (4-layer MLPs for both the node and edge encoder). The only difference is that its inputs do not contain any object appearance features.
    
    \item  \Btwo, a model that uses the 3D object centroid (X, Y, Z) and object-centric 2D image feature embeddings for forecasting 3D translation. This baseline model extracts 2D CNN features from each image, concatenates the features with the camera viewpoint, and transforms the combined features into an object appearance feature vector. The feature vector is concatenated with the 3D object centroid and fed into a graph network (identical to the one used in \Bone) to predict future object 3D translation. When taking multiple views as inputs, the model takes the average of the appearance feature vectors across views. 

    \item 
    Visual Foresight (\textit{VF}) \cite{ebert2018visual}, a model that uses the current frame and the action of the agent to predict future 2D frames by ``moving pixels" based on predicted 2D pixel flow fields. It is based on the publicly available code of  \cite{ebert2018visual} that uses such frame predictive model to infer an action trajectory that  brings an object pixel to the  desired (2D) location in the image space.
   
    \item 
    \Pla \cite{hafner2019planet}, a model that learns a scene-level embedding by predicting future frames and the reward given the current frame. 
    \Pla \ only deals with single-goal tasks and does not apply to our multi-goal pushing task. We extend it to our setting by appending the goal state to the observation. In practice, we augment the latent state vector produced from its state encoder’s first fully connected layer with a randomly selected goal, and provide the model with reward computed correspondingly. The reward at each timestep is the computed as the negative of the distance-to-goal.

    \end{enumerate} 
Note that both \textit{VF} and \Pla \ are self-supervised models that do not require ground-truth object states during training. However, we believe that since such supervision is readily accessible in simulation, we should leverage them to push the performance of the learned dynamics model. Self-supervised models are more favored when trained directly in the real world, where strong supervisions are not available, but as we showed in our experiments, our model trained solely in simulation can transfer reasonably well to the real world without any fine-tuning. As a result, we believe including the comparison with such self-supervised baselines is arguably fair and reasonable.

\subsection{Details for Pushing with MPC}
Here we described details of pushing with MPC discussed in section 4.3.
\newline
\subsubsection{Pushing in simulation}
\noindent\textbf{Pushing without obstacle} We test the performance of our model with MPC to push objects to desired locations. We run 50 experiments in the Bullet simulator. For each testing sample, we place either 1 or 2 objects in the $0.6m \times 0.6m$ workspace randomly, and sample a random goal for each object. The maximum distance of the goal to the initial position for each object is capped at $0.25m$.  For our model and \Btwo, we use a single randomly sampled view. For \VF \ and \Pla, we use a fixed top-down view for both training and testing. We set the maximum number of steps for each action sequence to be 10, and evaluate 30 random action sequences before taking an action. We use planning horizon of 1 since greedy action selection suffices for this task. The results are reported in Table 3 in the main paper. Note that we also train and test variants of VF and PlaNet to take observations from varying camera viewpoints, together with camera pose information. However, they both fail completely on this task.
\newline

\noindent\textbf{Collision-free pushing}
In order to test our model’s multi-step prediction performance, we evaluate our model on pushing in scenes with randomly sampled obstacles, and the robot is required to push an object to desired goal without colliding into any obstacle. For quantitative evaluation, we randomly place an object of interest and a goal position in the planar workspace. One obstacle object is placed between them with a small perturbation, so that there exists no straight collision-free path to reach the goal. The distance from the object to its goal is uniformly sampled from the range $[0.24m, 0.40m]$. Similarly, we run 50 examples, and use only one randomly selected camera view as input to our model. We evaluate 60 randomly sampled action sequences with length of 25 steps, and use a planning horizon of 10 steps. We achieve a success rate of \textbf{0.68} for this task.

Since randomly placing \textit{multiple} obstacles in the scene for quantitative evaluation while ensuring existence of collision-free path is non-trivial, we show qualitative planning results for such complex scenes in the supplementary video.

For both with- and without- obstacle pushing, it is considered a successful pushing sequence if all objects end up within 4cm (about half of the average object size) from the target positions on average. 

\begin{figure}[ht!]
\centering
\begin{minipage}{.5\textwidth}
  \centering
  \includegraphics[width=.9\linewidth]{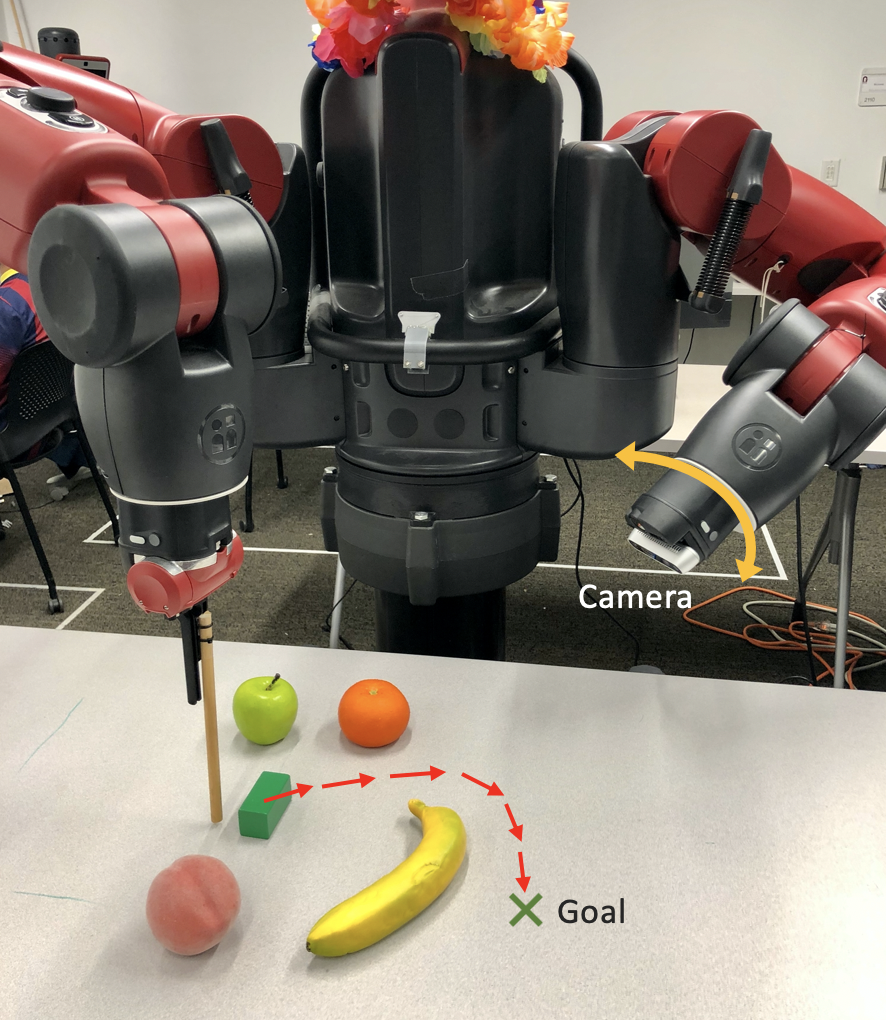}
  \caption{Real-world setup with Baxter}
  \label{fig:baxter}
\end{minipage}%
\begin{minipage}{.5\textwidth}
  \centering
  \includegraphics[width=.9\linewidth]{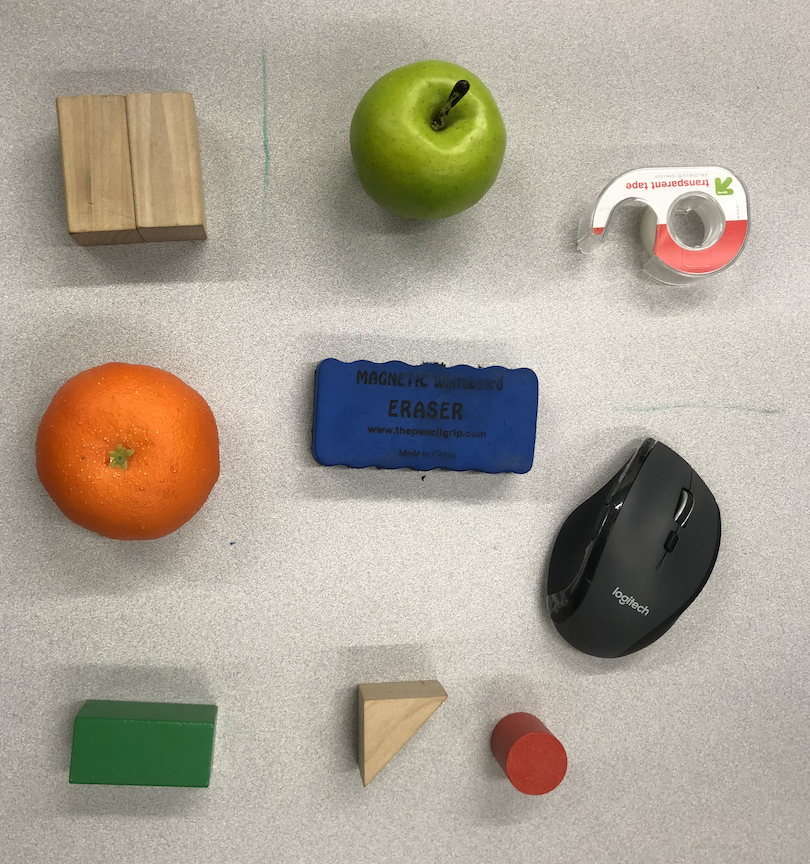}
  \caption{Objects for real-world experiments}
  \label{fig:obj}
\end{minipage}
\end{figure}

\subsubsection{Sim-to-real transfer for pushing in the real world}

We use a Baxter robot equipped with a rod-shaped end-effector attached to its right hand, similar to the setting in the Bullet simulation. One Intel RealSense D435 RGB-D camera is attached to the robot's left hand, and we use only one view for our experiment, as shown in Figure \ref{fig:baxter}. Please refer to the supplementary video for more qualitative results.

Due to reachability considerations, we down-scaled the size of the planar workspace by twice from the one in simulation, resulting a workspace of $0.3m\times0.3m$. For a fair comparison, we also down-scaled with the same factor the object-to-goal distance, length of horizontal movement per action step, and size of the tolerance for determining success/failure. We pick 20 objects with size of  $5$ to $10 cm$, which are commonly seen in a office setting, including fruits, wooden blocks, and stationery, and evaluate 5 pushing samples for each of them. Some of objects selected are shown in Figure \ref{fig:obj}.

For object detection in the real-world, we train our 3D detector using simulated data, and fine-tune it using a small set of real data (100 images capturing 25 distinct object configurations) collected using 4 cameras. The ground truth bounding-boxes and segmentation masks are obtained via background subtraction.

\end{document}